\documentclass{article}

\usepackage[final]{neurips_2022}

\usepackage[utf8]{inputenc} 
\usepackage[T1]{fontenc}    
\usepackage{hyperref}       
\usepackage{url}            
\usepackage{booktabs}       
\usepackage{amsfonts}       
\usepackage{nicefrac}       
\usepackage{microtype}      
\usepackage{xcolor}         

\usepackage{array}
\usepackage{graphicx}
\usepackage{subfigure}
\usepackage{multirow}

\RequirePackage{algorithm}
\RequirePackage{algorithmic}
\usepackage{float}

\title{On the Robustness and Anomaly Detection of Sparse Neural Networks}

\author{%
  Morgane Ayle \\ Technical University of Munich \\ morgane.ayle@tum.de\And Bertrand Charpentier \\ Technical University of Munich \\ bertrand.charpentier@in.tum.de \And John Rachwan \\ Technical University of Munich \\ john.rachwan@tum.de \And Daniel Zügner \\ Technical University of Munich \\ zuegnerd@in.tum.de \And Simon Geisler \\ Technical University of Munich \\ simon.geisler@in.tum.de \And Stephan Günnemann \\ Technical University of Munich \\ guennemann@in.tum.de
}

\begin{document}

\maketitle

\begin{abstract}
The robustness and anomaly detection capability of neural networks are crucial topics for their safe adoption in the real-world. Moreover, the over-parameterization of recent networks comes with high computational costs and raises questions about its influence on robustness and anomaly detection. In this work, we show that sparsity can make networks more robust and better anomaly detectors. To motivate this even further, we show that a pre-trained neural network contains, within its parameter space, sparse subnetworks that are better at these tasks without any further training. We also show that structured sparsity greatly helps in reducing the complexity of expensive robustness and detection methods, while maintaining or even improving their results on these tasks. Finally, we introduce a new method, SensNorm, which uses the sensitivity of weights derived from an appropriate pruning method to detect anomalous samples in the input.
\end{abstract}
\section{Introduction}
\looseness=-1
With the need to deploy Machine Learning (ML) models in real-life systems such as autonomous vehicles, medical diagnosis, online fraud detection etc., building safe and reliable ML pipelines has become more critical than ever. However, there exist three well-known cases of changes to the in-distribution data -- or anomalies -- to which models are known to be brittle: adversarial attacks (AA), which are intentionally crafted but imperceptible perturbations that can fool a classifier \cite{goodfellow2014explaining}; distribution shifts (DS), which are naturally occurring perturbations that can also change the model’s output \cite{hendrycks2019benchmarking}; and out-of-distribution (OOD) samples which come from an entirely different distribution, and that can surprisingly lead to highly confident but wrong predictions \cite{hendrycks2016baseline}. This latter case is notoriously known for generative models, which tend to assign higher likelihoods to samples coming from a distribution that they have not been trained on \cite{nalisnick2018deep}.

\looseness=-1
In an orthogonal line of work, Neural Networks (NN) are also known to be highly over-parameterized, which is made evident by the existence of sparse subnetworks that achieve the same accuracy performance as their dense counter-part \cite{han2015deep}. To find these subnetworks, many works have derived pruning criteria that assign scores to weights, and prune those with the lowest values \cite{lee2018snip, Wang2020picking}. However, the algorithms that attempt to find these subnetworks rarely evaluate them on other challenging tasks such as robustness and anomaly detection.

\looseness=-1
In this paper, we show that pruning can benefit robustness and anomaly detection by tackling the following four questions: 1) How does sparsity affect various properties of networks, beyond test accuracy? (Sec. \ref{section:properties}), 2) Can we find sparse subnetworks within pre-trained dense networks that achieve better robustness and detection? (Sec. \ref{section:finding}), 3) Can sparsity benefit tasks beyond deterministic single-network classification such as ensembles, Bayesian NN and generative models? (Sec. \ref{section:beyond}), and 4) Are pruning criteria useful for the detection of anomalies in the input? (Sec. \ref{section:detecting}) \footnote{Project page including code at https://www.cs.cit.tum.de/daml/uncertainty-snn/}
\section{Related Work}
\label{section:related}
\looseness=-1
\textbf{Sparse Networks.}
Pruning can be performed either before training a model \cite{lee2018snip, Wang2020picking}, early in training \cite{rachwan2022winning, You2020Drawing}, or after the model has been fully trained \cite{frankle2020linear, frankle2020stabilizing, lecun, gohil2020one}. 
It can be done either in an unstructured \cite{nalisnick2018deep, Wang2020picking, frankle2020linear} or a structured manner \cite{rachwan2022winning, li2016pruning, you2019gate}, where unstructured pruning removes weights by setting them to 0 whereas structured pruning removes entire neurons or filters by reducing the size of the weight matrices involved. Additionally, pruning can be done globally \cite{rachwan2022winning, lee2018snip, Wang2020picking, frankle2020linear} or locally \cite{ramanujan2020s}, where global pruning automatically decides how much to prune from each layer when given a desired total sparsity, whereas local pruning prunes the same percentage of weights at every layer. Unless otherwise specified, we use global pruning throughout the paper. We briefly introduce the 8 pruning algorithms we use in this paper:

\looseness=-1
\textit{\textbf{Pruning Before Training.}}
    \underline{SNIP \cite{lee2018snip}}: A sensitivity-based pruning method (i.e. it scores weights based on their sensitivity to a certain metric) that aims at preserving the loss function. The score used is: \smash{$s_i = |w_i \frac{\partial L}{\partial w_i}|$}, where $L$ is the loss and $w_i$ the weight.\\
    \underline{GRASP \cite{Wang2020picking}}: A sensitivity-based pruning method that aims at increasing the gradient flow. The final score is: \smash{$s_i = -|w_i(\textbf{H}\frac{\partial L}{\partial \textbf{w}})_i|$}, where \textbf{H} denotes the Hessian.\\
    \underline{CroPit \cite{rachwan2022winning}}: A sensitivity-based pruning method that aims at preserving the gradient flow. The final score is: $s_i = |w_i(\textbf{H}\frac{\partial L}{\partial \textbf{w}})_i|$, where \textbf{H} denotes the Hessian. We also use its structured version CroPit-S.\\
    \underline{Edge-popup \cite{ramanujan2020s}}: An optimization-based method that aims at finding sparse network by optimizing the mask directly instead of the weights. The sparse model is not trained further.
    
\looseness=-1
\textit{\textbf{Pruning During Training.}}
    \underline{IMP \cite{frankle2020linear}}: An iterative-based pruning method that prunes the smallest weights first. It goes through multiple train-prune-rewind cycles until it reaches its desired sparsity, where "rewind" corresponds to resetting the values of the remaining weights after each pruning phase to their value at an early point in training.\\
    \underline{EarlyCroP}: CroPit performed early in training instead of before training.

\looseness=-1
\textit{\textbf{Pruning After Training.}}
    \underline{SNIP-After}: SNIP applied on a fully trained network.\\
    \underline{Edge-popup-After}: Edge-popup applied on a fully trained network instead of randomly initialized one.

\looseness=-1
\textbf{Robustness and Anomaly Detection of Networks.} While NN are constantly improving in terms of accuracy or generative capability, their weak robustness and anomaly detection pose challenges to their safe deployment in the real world \cite{shafaei2018less}. Classifiers have shown to be un-calibrated, i.e. assign high confidence to wrong predictions, and works have tried to improve existing models \cite{hendrycks2018deep, liu2020simple, huang2021importance} or come up with new methods that focus solely on the detection of anomalous inputs \cite{tack2020csi}. Generative models that can compute the log-likelihood of inputs have also been shown to assign \textit{higher} likelihood to \textit{anomalous} samples \cite{nalisnick2018deep}, leading to another body of works that attempts to understand and solve this problem \cite{schirrmeister2020understanding}. Note that, unless otherwise specified, we use the common baseline Maximum-Softmax-Probability (MSP) \cite{hendrycks2016baseline} to compute the AUC-ROC scores throughout the paper, which uses the predicted probability of an input as the confidence.

\looseness=-1
\textbf{Robustness and Anomaly Detection of Sparse Networks.}
With the growing interest in sparsity of deep NN, the literature started addressing the question of whether sparse models would have different properties than their dense counterparts, beyond their test accuracy. 
Some works argue that an appropriate level of sparsity can improve robustness of models either to adversarial attacks \cite{guo2018sparse} or to distribution shifts \cite{diffenderfer2021winning}, while others argue the contrary \cite{verdenius2020pruning, liebenwein2021lost}. Another body of works focuses on simultaneously optimizing for high accuracy, robustness and sparsity by incorporating objectives that promote the latter two during training \cite{madaan2020adversarial, zhang2021can, sehwag2020hydra}. In contrast, we do not directly optimize weights for the tasks of robustness and detection, but evaluate the effect of existing sparsity methods on these tasks and propose new ways to take advantage of them.
\section{Properties of Sparse Classification Networks}
\label{section:properties}
\looseness=-1
We start by addressing the question: \textit{How does sparsity affect various properties of networks, beyond test accuracy?}
Our goal is to give a comprehensive view on the effect of pruning on robustness to and detection of anomalies. We therefore evaluate models pruned with 8 different pruning algorithms, and make the distinction between pruning before, during or after training. Indeed, we expect the pruning time to affect the set of weights being preserved, and therefore their final performance on new tasks.

\looseness=-1
\textbf{Setup.} We train ResNet18 models on the CIFAR10 dataset. We use L-$\infty$ and L-2 FGSM attacks with $\epsilon=8$, CIFAR10C as the DS dataset, and SVHN and CIFAR100 as the OOD datasets. We report relative accuracy (i.e. normalized with respect to the dense model) and relative AUC-ROC on attacks and shifts, and relative Brier score and relative AUC-ROC on OOD datasets. All models are trained with 3 different seeds and their average performance is reported. 

\begin{figure*}[h]
    \centering
    \includegraphics[width=\textwidth]{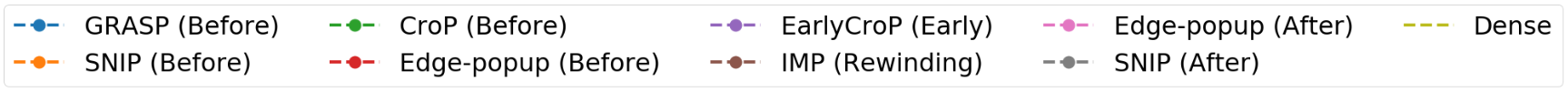}
        \subfigure[]{\includegraphics[width=0.3\textwidth]{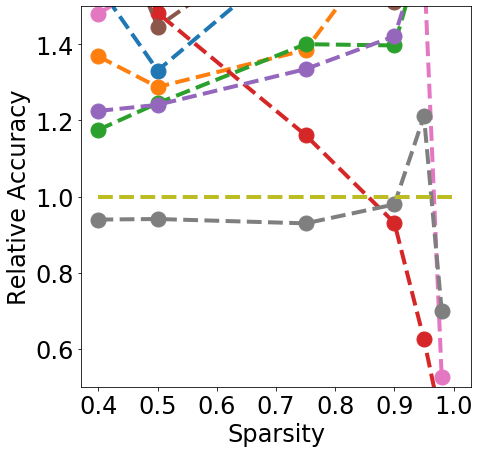}} 
    \subfigure[]{\includegraphics[width=0.3\textwidth]{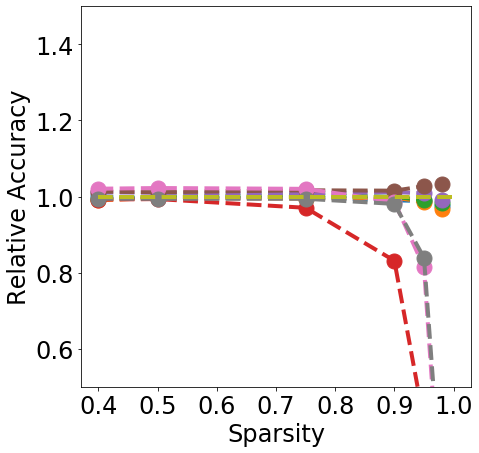}} 
    \subfigure[]{\includegraphics[width=0.3\textwidth]{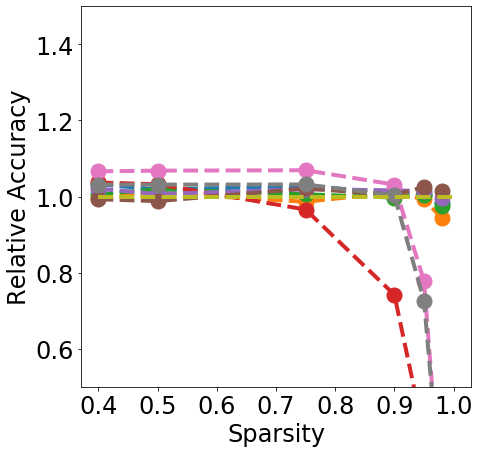}} 
    \caption{Comparison between the relative accuracy (higher is better) of a dense ResNet18 model and sparse ResNet18 models on (a) L-$\infty$ FGSM adversarial attacks with $\epsilon=8$, (b) L-2 FGSM adversarial attacks with $\epsilon=8$ and (c) CIFAR10C corruptions.}
    \label{fig:properties_robustness}
\end{figure*}

\begin{figure*}
    \centering
    \includegraphics[width=\textwidth]{fig/legend.png}
    \subfigure[]{\includegraphics[width=0.3\textwidth]{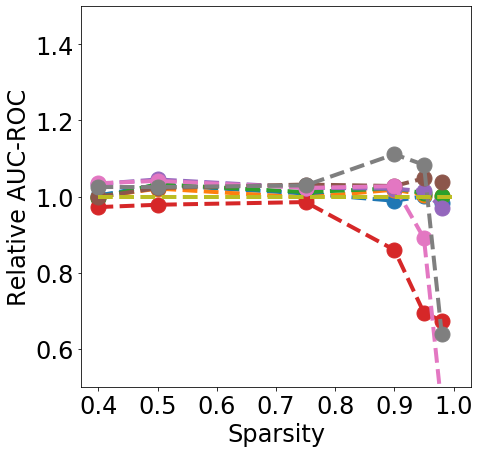}} 
    \subfigure[]{\includegraphics[width=0.3\textwidth]{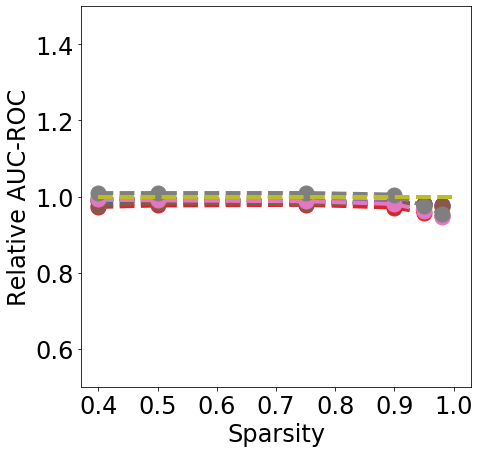}}
    \subfigure[]{\includegraphics[width=0.3\textwidth]{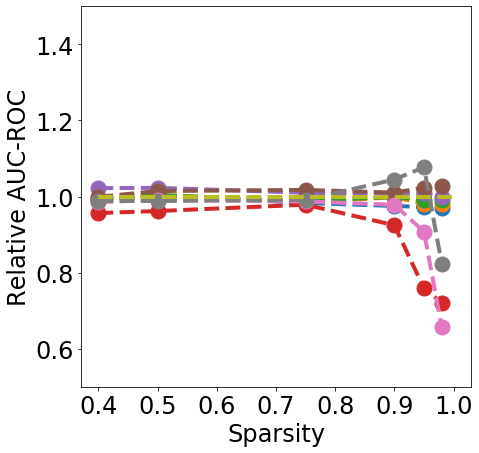}}
    \caption{Comparison between the relative AUC-ROC (higher is better) of a dense ResNet18 model and sparse ResNet18 models on (a) L-$\infty$ FGSM adversarial attacks with $\epsilon=8$, (b) L-2 FGSM adversarial attacks with $\epsilon=8$ and (c) CIFAR10C corruptions.}
    \label{fig:properties_detection}
\end{figure*}

\begin{figure*}
    \centering
    \includegraphics[width=\textwidth]{fig/legend.png}
    \subfigure[]{\includegraphics[width=0.3\textwidth]{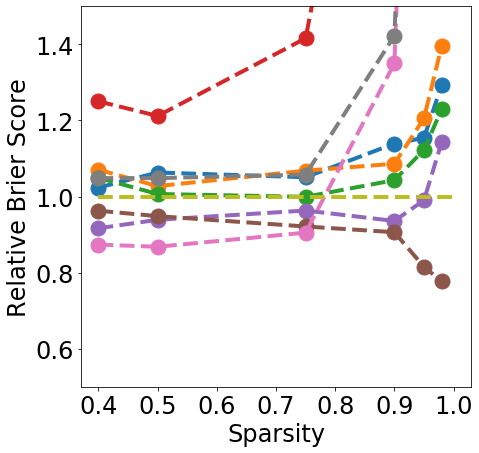}}
    \subfigure[]{\includegraphics[width=0.3\textwidth]{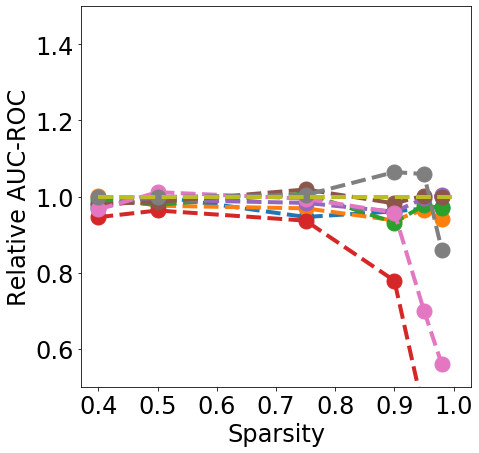}}
    \subfigure[]{\includegraphics[width=0.3\textwidth]{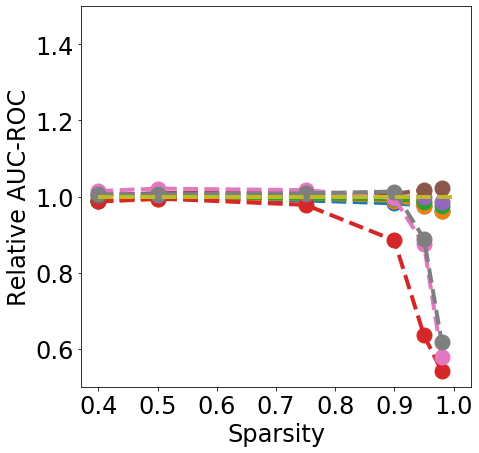}} 
    \caption{Comparison between a dense ResNet18 and sparse ResNet18 models of (a) the relative Brier score (lower is better), and the relative AUC-ROC on (b) SVHN and (c) CIFAR100.}
    \label{fig:properties_uncertainty}
\end{figure*}

\textbf{Discussion.}
\looseness=-1
Fig. \ref{fig:properties_robustness} clearly shows that almost all pruning methods improve the robustness to L-$\infty$ FGSM attacks (a), but not to L-2 FGSM attacks (b). Indeed, the relation between L-$\infty$ and sparsity has already been discussed in \cite{guo2018sparse}, and works such as \cite{verdenius2020pruning} have reported a \textit{decrease} of robustness on L-2 based attacks such as Carlini-Wagner \cite{carlini2017towards}. We also observe that most methods \textit{maintain} the robustness to DS (c), with During-Training methods performing slightly better than Before-Training methods at high sparsity.

\looseness=-1
In terms of detection, Fig. \ref{fig:properties_detection} shows that generally all methods except Edge-popup can maintain the AUC-ROC on DS. SNIP-After also appears to provide better improvements at an appropriate sparsity than the remaining methods.

\looseness=-1
Finally, from Fig. \ref{fig:properties_uncertainty} we can see that During or After-Training pruning methods perform better than the dense model in terms of Brier score (a) up until 90\% sparsity, but only IMP does so at even higher sparsity. Generally all methods, except Edge-popup, can also maintain the AUC-ROC on the OOD datasets SVHN (b) and CIFAR100 (c). We report additional metrics and results on LeNet5 in App. \ref{appendix:properties}.

\section{Finding robust subnetworks within non-robust dense networks}
\label{section:finding}
\looseness=-1
Next, we address the question: \textit{Can we find  (sparse) sub-networks within dense pre-trained networks that achieve better robustness and detection?} This question has been briefly tackled by \cite{sehwag2020hydra}, but they only evaluated the robustness of the sparse models to adversarial attacks, while we aim to evaluate their robustness to both adversarial attacks and distribution shifts, as well as their OOD detection capability.

\looseness=-1
\textbf{Setup.} We train ResNet18 models on CIFAR10, and report the accuracy and AUC-ROC on FGSM attacks with $\epsilon=8$ and on CIFAR10C, as well as the AUC-ROC on SVHN. Similarly to \cite{sehwag2020hydra}, we use the Edge-popup algorithm, which trains a \textit{mask over the weights} instead of the weights themselves (App. \ref{app:algorithms}). However instead of using the standard cross-entropy objective $\mathcal{H}$, we augment the objective depending on the task we are trying to solve. In the following, $X$ refers to a clean batch, $Y$ to its ground-truth labels, and $f$ to the model:\\
\looseness=-1
\underline{Adversarial Attacks}: $\mathcal{L}(X) = \mathcal{H}(f(X), Y) + \lambda * \mathcal{KL}(f(X), f(X'))$, with $X$with $X'$ the adversarial attack of $X$, and $\mathcal{KL}$ the KL divergence between the predictions on $X$ and $X'$. The second term in $\mathcal{L}$ ensures that the predictions on $X$ and $X'$ do not deviate too much. We use $\lambda = 6$.\\
\underline{Distribution Shifts}: $\mathcal{L}(X) = \mathcal{H}(f(X'), Y)$, with $X'$ being a clean batch perturbed with gaussian noise. Indeed, training a model on clean inputs perturbed with gaussian noise can lead to a model more robust to natural perturbations \cite{diffenderfer2021winning}.\\
\underline{OOD Detection}: $\mathcal{L}(X) = \mathcal{H}(f(X), Y) + \mathcal{H}(f(X'), U)$, with $X'$ being an OOD batch and $U$ a matrix of uniform distributions. The second term in $\mathcal{L}$ ensures the predictions on OOD samples are as close as possible to $U$ (maximum uncertainty).

\begin{table*}[h!]
\centering
\small
\begin{tabular}{m{3.7cm}|m{1.3cm}|m{1.4cm}|m{1.4cm}|m{1.5cm}|m{1.5cm}} 
 \toprule
  & Accuracy FGSM 8 & AUC-ROC FGSM 8 & AUC-ROC SVHN & Accuracy CIFAR10C & AUC-ROC CIFAR10C \\ 
 \midrule
 Dense & 19.9\% & \textbf{79.5\%} & 90.3\% & 66.2\% & \textbf{68.4\%} \\ 
 \midrule
 Edge-popup - AA Objective & \textbf{72.9\%} & 53.8\% & 77.6\% & 68.8\% & 57.3\% \\ 
 \midrule
 Edge-popup - OOD Objective & 60.2\% & 64.7\% & \textbf{99.7\%} & 64.5\% & 67.3\% \\
 \midrule
 Edge-popup - DS Objective & 64.4\% & 56.7\% & 78.0\% & \textbf{70.0\%} &  54.5\% \\
\bottomrule
\end{tabular}
\caption{Accuracy and AUC-ROC of ResNet18 models pruned with Edge-popup to 50\% sparsity based on AA, OOD or DS objectives. }
\label{table:finding}
\end{table*}

\looseness=-1
\textbf{Discussion.} In Tab. \ref{table:finding}, we report the metrics of interest for the three objectives on Edge-popup and compare to the original dense model. Each objective improves its corresponding metric the most, but sometimes also benefits others. For example, training on the OOD objective significantly improves the accuracy on FGSM attacks, a correlation which has already been noted in the literature \cite{lee2020removing}. This experiment shows that subnetworks that achieve much better performance on a certain objective do exist wihtin a pre-trained dense network. This suggests that dense networks focus on non-robust features as was shown in \cite{ilyas2019adversarial}, possibly due to their over-parameterization, and pruning eliminates some weights that focus on these features.
\section{The Effect of Sparsity Beyond Single-Network Classification}
\label{section:beyond}
\looseness=-1
In this section, we investigate the effect of sparsity on existing methods that improve the robustness or detection capability of networks, and try to answer the question: \textit{Can sparsity benefit tasks beyond deterministic single-network classification?} We examine ensembles, Bayesian neural networks and generative models.

\textbf{Ensembles.}
\looseness=-1
One popular method to improve robustness and calibration of classifiers is ensembles. It consists of training multiple networks independently, and then averaging their outputs to obtain more calibrated predictions \cite{lakshminarayanan2016simple}. Ensembles have been shown to be among the most calibrated methods against distribution shifts \cite{ovadia2019can}. However, one major drawback is the expensive nature of this method, both in terms of memory and time. Structured pruning helps in overcoming both of these drawbacks, since it removes entire neurons and channels from the architecture, thereby reducing the training and inference time, but also the space complexity. For this experiment, we use the large model AlexNet trained on CIFAR10, and compare a single dense model and an ensemble of 5 dense models to an ensemble of 5 models pruned to 80\% node sparsity each. We can see from Tab. \ref{table:ensembles} that the sparse ensemble reaches almost the same test accuracy as the dense ensemble, matches it in terms of AUC-ROC on SVHN, and is much more efficient than the dense ensemble in terms of training and inference time, and GPU RAM usage.

\begin{table*}[h!]
\centering
\small
\begin{tabular}{m{2.8cm}|m{1.5cm}|m{1.5cm}|m{1.3cm}|m{1.3cm}|m{0.8cm}} 
 \toprule
  & Test \newline Accuracy & AUC-ROC SVHN & Training Time & Inference Time & RAM \\ 
 \midrule
 Dense - Deterministic & 85\% & 83\% & x1 & x1 & x1 \\ 
 \midrule
 Dense - Ensemble & \textbf{90\%} & \textbf{89\%} & x5 & x4.8 & x5 \\ 
 \midrule
 CroPit-S - Ensemble & 88.4\% & \textbf{89\%} & \textbf{x2.7} & \textbf{x3.8} & \textbf{x0.42} \\
\bottomrule
\end{tabular}
\caption{Comparison between a dense AlexNet model, an ensemble of 5 dense AlexNet models annd an ensemble of 5 AlexNet models pruned with CroPit-S to 80\% node sparsity trained on CIFAR10.}
\label{table:ensembles}
\end{table*}

\textbf{Bayesian Neural Networks.}
\looseness=-1
Bayesian NN also fall into the category of expensive uncertainty methods. They are both slower to train and infer on, and more memory intensive since they often require to learn more parameters than a dense model. To illustrate the potential of sparsity in such tasks, we turn our attention to the popular method presented in \cite{blundell2015weight}. It consists of learning both a mean $w_i$ and a standard deviation $t_i$ for each weight $i$ in the network. At test time, we sample a value for each weight when doing an inference, and repeat the process multiple times to obtain multiple predictions, similar to ensembles. This work suggests to remove weights with a low Signal-to-Noise Ratio (SNR) $s_i = |\frac{w_i}{t_i}|$, but only evaluates the pruned model on the final accuracy. SNR is an intuitive pruning criterion, since it encourages the removal of weights with small means (similar to IMP which removes weights close to 0), and high standard deviation which indicates the uncertainty of the model and therefore the vulnerability of this weight to changes in the input. We extend this criterion to a structured one by simply summing the scores at each activation, which we call SNR-S, to take advantage of the removal of structures to achieve faster training and inference time. We train a dense Conv6 model on CIFAR10, a deterministic one, and two SNR-S models where pruning is done either during or after training. From Tab. \ref{table:bayesian}, we observe that both versions of Bayesian SNR-S achieve better accuracy than the dense Bayesian, and SNR-S After achieves even better AUC-ROC than both dense models, while being more efficient than the dense Bayesian model.

\begin{table*}[h]
\centering
\small
\begin{tabular}{m{3.4cm}|m{1cm}|m{1cm}|m{1.1cm}} 
 \toprule
  & Test Accuracy & AUROC SVHN & Inference Time \\ 
 \midrule
 Dense - Deterministic & \textbf{0.874} & 0.834 & \textbf{x1} \\ 
 \midrule
 Dense - Bayesian & 0.864 & 0.873 & x2.6 \\ 
 \midrule
 SNR-S (After) - Bayesian & 0.865 & \textbf{0.883} & \underline{x1.6} \\
 \midrule
 SNR-S (During) - Bayesian & \underline{0.871} & \underline{0.882} & \underline{x1.6}  \\
\bottomrule
\end{tabular}
\caption{Comparison of Conv6 trained on CIFAR10 in a deterministic, Bayesian or sparse Bayesian setting. For the Bayesian results we use 5 inferences and the sparse models are pruned to 50\% node sparsity.}
\label{table:bayesian}
\end{table*}

\textbf{Generative Models.}
\looseness=-1
Sparsity in generative models has recently started to gain attention in the literature \cite{kalibhat2020winning}. However, to the best of our knowledge, its effect on OOD detection has not been investigated yet. Generative models, some of which are capable of computing the log-likelihood of input samples, are known for their counter-intuitive behavior on OOD samples: they assign \textit{higher} likelihood to OOD samples than to their own training/testing in-distribution samples. 
\looseness=-1
In this experiment, starting with the pre-trained generative model Glow \cite{kingma2018glow}, we prune it using EarlyCroP \textit{locally}. We observe that, in the FashionMNIST-MNIST combination, pruning greatly helps in better distinguishing the two types of datasets and correctly assign \textit{lower} likelihoods to OOD samples. However, when moving to more complex cases (CIFAR10-SVHN or CELEBA-SVHN), we do not observe the same behavior. Nevertheless, the success on the FashionMNIST-MNIST combination encourages further exploration of the benefits of sparsity in generative models. Note that local pruning plays a crucial role here, as global pruning did not result in the same behavior on FashionMNIST. One possible explanation for this is that local pruning creates more bottlenecks in the network, which has been shown to help in OOD detection \cite{kirichenko2020normalizing}.

\begin{figure}[h]
    \centering
    \subfigure[]{\includegraphics[width=0.23\textwidth]{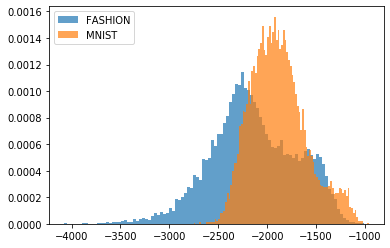}} 
    \subfigure[]{\includegraphics[width=0.23\textwidth]{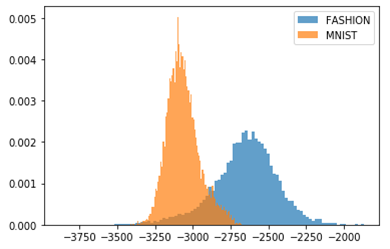}} 
    \hfill
    \subfigure[]{\includegraphics[width=0.23\textwidth]{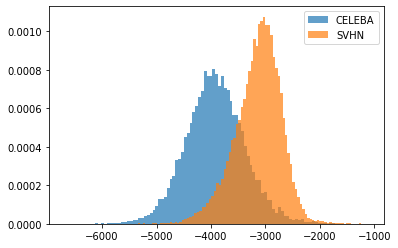}} 
    \subfigure[]{\includegraphics[width=0.23\textwidth]{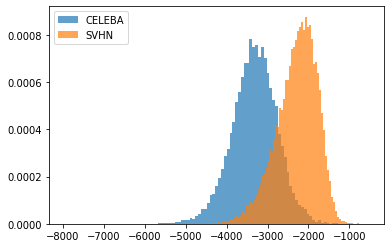}} 
    \caption{Left: Count vs input log-density of Glow trained on FashionMNIST, tested on the OOD dataset MNIST with (a) no pruning and (b) local pruning to 90\% using EarlyCroP. Right: Count vs input log-density of Glow trained on CELEBA, tested on the OOD dataset SVHN with (c) no pruning and (d) local pruning to 90\% using EarlyCroP.}
    \label{fig:glow_fashion}
\end{figure}

    
\section{SensNorm - Detecting Anomalies Through Weight Sensitivity}
\label{section:detecting}
\looseness=-1
We finally address the question: \textit{Are pruning criteria useful for the detection of anomalies in the inputs?}
Some pruning methods that operate \textit{before} training have been shown to be independent of the batch being used to score the weights \cite{su2020sanity}. This means that the weights being pruned and the final structure of the network do not depend much on the current task/dataset of interest. However, methods operating \textit{during} training are believed to use information from the batch to keep the most important weights for the given task \cite{su2020sanity}. Note that this applies to methods that use a sensitivity criterion, i.e. that assign a score to each weight (unlike IMP). Since pruning based on one or more batches will preserve the weights that are relevant for these batches, pruning based on batches that contain anomalies will attempt to preserve a different set of weights. Based on these observations, our key idea is to detect anomalies by measuring the distance between the sensitivity of weights on in-distribution (training) data, and that of each incoming test batch. Intuitively, the larger the distance the more likely the input batch contains anomalous samples.

\looseness=-1
\textbf{Method.} More formally, we are given a pre-trained model with weights $\mathbf{w}$, a train dataset \smash{$D_{train}=\{d_i\}_{i=0}^{m}$}, a test dataset \smash{$D_{test}=\{d'_j\}_{j=0}^{n}$}, and a sensitivity scoring function $f_{B}$ where $B$ indicates one or more batches used to get the weights' sensitivity. We first compute the sensitivity of weights on the training data $f_{D_{train}}(\mathbf{w})$ by going through all training batches and accumulating the gradients. We also save the element-wise standard deviation over all batches $std(f_{D_{train}}(\mathbf{w}_i))$. At test time, for every incoming batch $d'_j$, we compute $f_{d'_j}(\mathbf{w})$. Finally, we compute the distance \smash{$S_{d'_j} = \|f_{D_{train}}(\mathbf{w}) - f_{d'_j}(\mathbf{w})\|_p$}, where $\|.\|_p$ is the L$p$-norm (in practice we use $p=5$). If $S_{d'_j} \geq \tau$, where $\tau$ is some pre-defined threshold, then $d'_j$ is an anomalous batch; otherwise, it is an in-distribution batch. We refer to our method as SensNorm.

\looseness=-1
We rely on the SNIP score as our sensitivity score, since its purpose is to preserve the loss of the task at hand. However, we find that dropping the absolute value helps in achieving better performance. Indeed, the  sign of a score in SNIP indicates whether a certain weight was increasing or decreasing the loss, which can help distinguish in-distribution from anomalous batches. We also standardize the test batch scores by their mean and standard deviation computed on the training data, so as to reduce the contribution of the weights that are highly variable on the in-distribution data itself. Our weight sensitivity function is therefore:
    
\begin{equation}
f_B(\mathbf{w}_i)=\frac{{}\mathbf{w}_i \frac{\partial L}{\partial \mathbf{w}_i} - f_{D_{train}}(\mathbf{w}_i)}{std(f_{D_{train}}(\mathbf{w}_i))}
\end{equation}

\looseness=-1
\textbf{Discussion.} Our method naturally lends itself to batch detection, but can also be applied for single-input prediction. Tab. \ref{table:wi_batches} shows that, with 100 samples, our method can detect all types of inputs easily, and their detection remains high until a batch of size 5. The challenge arises for batches of size 2 and 1, since gradients will be very noisy for these cases. We therefore rely on augmentations to gain more information from the input sample (App. \ref{app:experimental}). Note that, while originally conceived for detecting batches, SensNorm still outperforms MSP on batch size 1. We also compare our results to GradNorm, which uses gradient norms to detect anomalies, and find that SensNorm matches it or even outperforms it on small batch sizes.

\begin{table*}[h!]
\centering
\resizebox{0.6\columnwidth}{!}{
\begin{tabular}{m{0.05cm}|m{1.3cm}|m{0.9cm}|m{1.3cm}|m{0.9cm}|m{0.8cm}|m{1cm}} 
 \toprule
 & & \small SVHN & \small CIFAR100 & \small FGSM & \small DS & \small OODom. \\
 \midrule
 \multirow{3}{*}{\rotatebox[origin=c]{90}{\tiny Batch size 100} } & \small MSP  & \textbf{100\%} & \textbf{100\%} & \textbf{100\%} & \underline{98.7\%} & \underline{51.3\%} \\
 \cmidrule{2-7}
   & \small GradNorm & \textbf{100\%} & \textbf{100\%} & \underline{99.8\%} & 91.9\% & 33.3\% \\
  \cmidrule{2-7}
  & \small SensNorm & \textbf{100\%} & \textbf{100\%} & \textbf{100\%} & \textbf{99.9\%} & \textbf{100\%}\\
   
 \midrule\midrule
 \multirow{3}{*}{\rotatebox[origin=c]{90}{\tiny Batch size 5  } } & \small MSP & 98.5\% & 92.7\% & 87.7\% & 84.5\% & \underline{50.8\%} \\
  \cmidrule{2-7}
  & \small GradNorm & \underline{99.8\%} & \underline{98.6\%} & \underline{94.9\%} & \underline{92.3\%} & 32.9\% \\
  \cmidrule{2-7}
  & \small SensNorm & \textbf{99.9\%} & \textbf{99.2\%} &\textbf{98.8\%} & \textbf{96.5\%} & \textbf{100\%} \\
 \midrule\midrule
 \multirow{3}{*}{\rotatebox[origin=c]{90}{\tiny Batch size 2} } & MSP & 92.2\% & 85.5\% & 81.2\% & 79.1\% & \underline{51.6\%} \\
 \cmidrule{2-7}
   & \small GradNorm & \textbf{98.8\%} & \textbf{92.9\%} & \textbf{88.3\%} & \underline{87.0\%} & 33.1\% \\
  \cmidrule{2-7}
   & \small SensNorm & \underline{97.3\%} & \underline{91.0\%} & \underline{87.7\%} & \textbf{87.3\%} & \textbf{100\%} \\
   \midrule\midrule
   \multirow{3}{*}{\rotatebox[origin=c]{90}{\tiny Batch size 1} }& \small MSP & 90.3\% & \underline{83.3\%} & 79.5\% & 77.1\% & \underline{53.4\%}\\
    \cmidrule{2-7}
    & \small GradNorm & \textbf{96.7\%} & \textbf{85.2\%} & \underline{80.4\%} & \textbf{81.0\%} & 33.0\%\\
    \cmidrule{2-7}
    & \small SensNorm & \underline{93.7\%} & \textbf{85.2\%} & \textbf{80.6\%} & \underline{79.0\%} & \textbf{100\%}\\
\bottomrule
\end{tabular}
}
\caption{AUC-ROC of SensNorm compared to MSP and GradNorm on multiple batch sizes and anomalies. Best result is in bold, second best result is underlined.}
\label{table:wi_batches}
\end{table*}

\section{Conclusion}
\looseness=-1
In conclusion, we show that the topics of pruning, robustness and detection are closely intertwined. First, most pruning algorithms can maintain the original uncertainty of the dense model on AA, DS and OOD. All pruning methods also greatly improve the accuracy to L-$\infty$ FGSM attacks, but not to the L-$2$ based attacks. Pruning methods that operate during training (i.e. EarlyCroP and IMP) often offer the best trade-off between good test performance and maintaining or even improving robustness and detection metrics. Second, we show that it is possible to find subnetworks within pre-trained dense networks that perform better than the original models in terms of robustness and detection. Third, we find that pruning can be beneficial in tasks such as ensembles, Bayesian NN and generative models, especially in terms of efficiency for the expensive uncertainty estimation methods. Finally, we introduce SensNorm, a new method for detecting batches of anomalous samples. SensNorm shows that weight sensitivity is a good indicator of the presence of anomalies in the input.

\newpage

\bibliography{main}
\bibliographystyle{icml2022}

\newpage

\appendix

\section{Experimental Setup}
\label{app:experimental}

\textbf{Architecture-Dataset Combinations}
For classification tasks, we focus on two popular architecture-dataset combinations: LeNet5 trained on MNIST, and ResNet18 trained on CIFAR10. For the Ensembles experiment, we use AlexNet trained on CIFAR10. For the Bayesian experiment, we use Conv6 trained on CIFAR10. For generative models, we use Glow trained on MNIST or CELEBA.

\textbf{Train-Test Dataset Combinations}
We evaluate the robustness of classification models on two types of dataset shifts:\\
\textit{Distribution Shifts (DS)}: we use the CIFAR10-C dataset on the ResNet18-CIFAR10 combination only.\\
\textit{Adversarial Attacks (AA)}: we use the Fast Gradient Sign Method (FGSM) with $\epsilon = 8$ in the [0, 255] scale.

We also evaluate the uncertainty of classification models on the previous two dataset types, and the two following additional dataset types for all models:\\
\textit{Out-of-Distribution (OOD)}: For FashionMNIST, we use MNIST. For CIFAR10, we use SVHN and CIFAR100. For CELEBA, we use SVHN. \\
\textit{Out-of-Domain (OODom)}: this corresponds to the any OOD dataset but scaled in the [0, 255] range instead of the [0, 1] range.

\textbf{Metrics} 
We report the clean test accuracy as well as the accuracy on DS and AA (Acc.), the Area Under the Receiver Operating Characteristic curve (AUROC) or Area Under the Precision Recall curve (AUPR) for detection, the Brier score as an uncertainty measure (Brier), and a lower bound on the Lipschitz constant of models (Lip.).

\textbf{Training Details}
Classifiers are trained using Adam optimizer, learning rate 2e-3  and batch size 512. ResNet18 and AlexNet are trained for 50 epochs, while LeNet5 is trained for 30. Glow is trained for 100k iterations, using a batch size of 64 and a learning rate of 1e-5.

\textbf{Agumentations used in SensNorm} We empirically find that rotation transformations in ${90, 180, 270}$, horizontal and vertical flips, as well as other affine transformations using \cite{benton2020learning}, help improve the detection.

\section{Algorithms}
\label{app:algorithms}

\begin{algorithm}[H]
Inputs: $\{X_n, Y_n\}_{n=1}^{N}$, $F$ pretrained network, x percentage to prune\\

Freeze F's weights\\
$S \leftarrow$ kaiming\_normal\\

\textbf{for} $n \in [1,N]$\\
\hspace*{3ex} Create subnetwork of (1-x)\% highest scores in S\\
\hspace*{3ex} Do forward pass\\
\hspace*{3ex} Compute loss $\mathcal{L}(x)$ with desired objective \hspace*{9ex}  \textit{\% Adversarial, OOD or DS}\\
\hspace*{3ex} Do backward pass through \textit{all} scores and update them\\

$F \leftarrow$ subnetwork of (1-x)\% highest scores in S\\
\caption{Finding Better Subnetworks within Dense Networks}
\label{method}
\end{algorithm}

\section{Additional Results}
\label{appendix:properties}

\begin{figure}[h]
    \centering
    \includegraphics[width=\textwidth]{fig/legend.png}
    \subfigure[]{\includegraphics[width=0.3\textwidth]{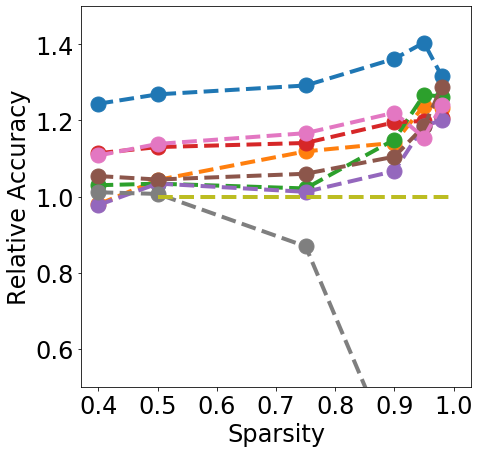}} 
    \subfigure[]{\includegraphics[width=0.3\textwidth]{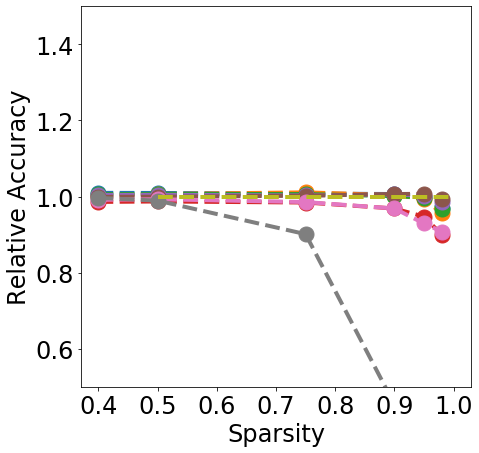}} 
    \caption{Robustness of the dense LeNet5 vs. pruned at different levels of sparsity to (a) L-$\infty$ FGSM adversarial attack with $\epsilon=8$, and (b) L-2 FGSM adversarial attack with $\epsilon=8$.}
    \label{fig:properties_robustness_lenet}
\end{figure}

\begin{figure}[h]
    \centering
    \includegraphics[width=\textwidth]{fig/legend.png}
    \subfigure[]{\includegraphics[width=0.3\textwidth]{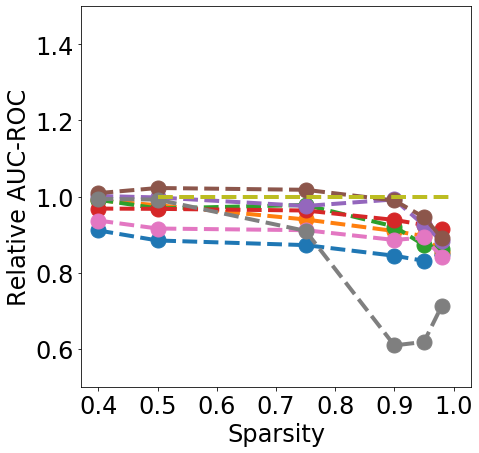}}
    \subfigure[]{\includegraphics[width=0.3\textwidth]{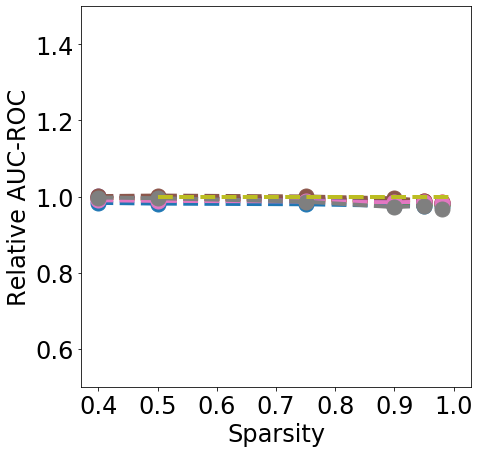}} 
    \caption{AUROC of the dense LeNet5 vs. pruned at different levels of sparsity to (a) L-$\infty$ FGSM adversarial attack with $\epsilon=8$, and (b) L-2 FGSM adversarial attack with $\epsilon=8$.}
    \label{fig:properties_detection_lenet}
\end{figure}

\begin{figure}[h]
    \centering
    \includegraphics[width=\textwidth]{fig/legend.png}
    \subfigure[]{\includegraphics[width=0.3\textwidth]{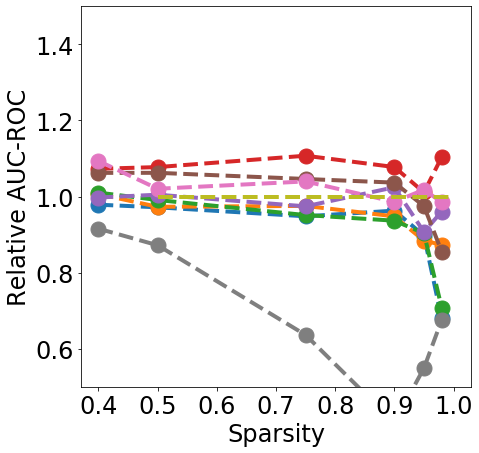}}
    \subfigure[]{\includegraphics[width=0.3\textwidth]{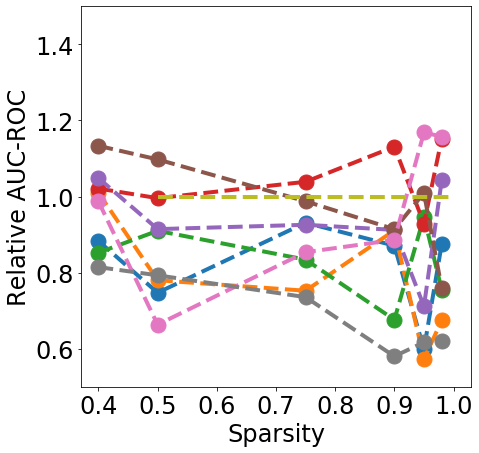}} 
    \caption{AUROC on (a) MNIST and (b) Omniglot of the dense LeNet5 vs. pruned at different levels of sparsity.}
    \label{fig:properties_detection_ood_lenet}
\end{figure}

We report additional information concerning the properties of sparse neural networks. Tab. \ref{table:lip} shows the Lipschitz constant \cite{scaman2018lipschitz} of dense and sparse ResNet18 models. Note that, at the exception of Edgepop, all pruning methods result in a \textit{higher} Lipschitz constant than the dense model. In Tab. \ref{table:cifar10c_5}, we look in details at the accuracy of pruned ResNet18 models on CIFAR10C, and whether or not it increased compared to the dense model. We note that the "saturate", "fog" and "contrast" corruptions are the most affected negatively among sparse models. Finally, Fig. \ref{fig:properties_robustness_lenet}, \ref{fig:properties_detection_lenet} and Fig. \ref{fig:properties_detection_ood_lenet} show the robustness, detection and uncertainty of AA, DS and OOD of LeNet5 models trained on MNIST. Performance on robustness and detection of FGSM attacks resembles the results on ResNet18. However, there is a much much noticeable improvement of certain pruning methods on the MNIST dataset, and a much more noticeable drop in detection of the Omniglot dataset. Indeed, the latter consists of a white background with black symbols, instead of black background with white digits for MNIST. 

\begin{table*}[h!]
\centering
\begin{tabular}{ m{3cm}|m{1.6cm}} 
 \toprule
  & Lipschitz Constant \\ 
 \midrule
 Dense & 7.27e+18 \\ 
 Edgepop (50\%) & 2.92e+11 \\
 CROP (50\%) & 6.62e+27 \\
 EarlyCROP (50\%) & 8.95e+20 \\
 \bottomrule
\end{tabular}
\caption{Lipschitz Constant of dense and pruned ResNet18 models trained on CIFAR10}
\label{table:lip}
\end{table*}

\begin{table}[h]
\centering
\tiny
\begin{tabular}{m{1.0cm}|m{0.3cm}|m{0.3cm}|m{0.3cm}|m{0.3cm}|m{0.3cm}|m{0.3cm}|m{0.3cm}|m{0.3cm}|m{0.3cm}|m{0.3cm}|m{0.3cm}|m{0.3cm}|m{0.3cm}|m{0.3cm}|m{0.3cm}|m{0.3cm}|m{0.3cm}|m{0.3cm}|m{0.3cm}}
\toprule
    \rotatebox[origin=c]{90}{criterion} &  \rotatebox[origin=c]{90}{gaussian noise} &  \rotatebox[origin=c]{90}{impulse noise} &  \rotatebox[origin=c]{90}{defocus blur} &  \rotatebox[origin=c]{90}{frosted glass blur} &  \rotatebox[origin=c]{90}{saturate} &  \rotatebox[origin=c]{90}{shot noise} &  \rotatebox[origin=c]{90}{elastic} &  \rotatebox[origin=c]{90}{snow} &  \rotatebox[origin=c]{90}{spatter} &  \rotatebox[origin=c]{90}{fog} &  \rotatebox[origin=c]{90}{frost} &  \rotatebox[origin=c]{90}{gaussian blur} &  \rotatebox[origin=c]{90}{speckle noise} &  \rotatebox[origin=c]{90}{brightness} &  \rotatebox[origin=c]{90}{pixelate} &  \rotatebox[origin=c]{90}{jpeg compression} &  \rotatebox[origin=c]{90}{contrast} &  \rotatebox[origin=c]{90}{motion blur} &  \rotatebox[origin=c]{90}{zoom blur} \\
\midrule
                    Dense &                        0.227 &                       0.243 &                      0.520 &                            0.360 &                  0.812 &                    0.264 &                 0.642 &              0.633 &                 0.707 &             0.700 &               0.521 &                       0.410 &                       0.313 &                    0.850 &                  0.491 &                          0.658 &                  0.292 &                     0.569 &                   0.576 \\
                    
                    GRASP &                        \textbf{0.356} &                       \textbf{0.367} &                      \textbf{0.585} &                            \textbf{0.523} &                  \textbf{0.814} &                    \textbf{0.403} &                 \textbf{0.712} &              \textbf{0.695} &                 \textbf{0.735} &             \textbf{0.673 }&               \textbf{0.611} &                       \textbf{0.436} &                       \textbf{0.447} &                    \textbf{0.849} &                  \textbf{0.520} &                          \textbf{0.721} &                  0.288 &                     \textbf{0.591} &                   \textbf{0.607} \\
                    SNIP &                       \textbf{0.360} &                       \textbf{0.316} &                      \textbf{0.526} &                            \textbf{0.473} &                  \textbf{0.804} &                    \textbf{0.431} &                 \textbf{0.677} &              \textbf{0.680} &                 \textbf{0.730} &             \textbf{0.667} &               \textbf{0.581} &                       0.387 &                       \textbf{0.464} &                    0.847 &                  \textbf{0.505} &                          \textbf{0.713} &                  0.277 &                    \textbf{0.584} &                   \textbf{0.579} \\
                    CROP &                        \textbf{0.324} &                       \textbf{0.305} &                      \textbf{0.597} &                            \textbf{0.489} &                  \textbf{0.821} &                    \textbf{0.372} &                 \textbf{0.713} &             \textbf{0.703} &                 \textbf{0.724} &             0.678 &               \textbf{0.598} &                       \textbf{0.469} &                       \textbf{0.409} &                    \textbf{0.855} &                  \textbf{0.513} &                          \textbf{0.701} &                  0.273 &                     \textbf{0.623} &                   \textbf{0.646} \\
                    Edgepop &                        \textbf{0.437} &                       \textbf{0.333} &                      0.515 &                            \textbf{0.475} &                  0.775 &                    \textbf{0.477} &                 \textbf{0.698} &              \textbf{0.686} &                 \textbf{0.747} &             0.576 &               \textbf{0.618} &                       \textbf{0.423} &                       \textbf{0.489} &                    0.832 &                  \textbf{0.494} &                          \textbf{0.730} &                  0.269 &                     \textbf{0.577} &                   \textbf{0.582} \\
                        EarlyCROP &                        \textbf{0.323} &                       \textbf{0.292} &                      \textbf{0.527} &                            \textbf{0.455} &                  \textbf{0.822} &                    \textbf{0.387} &                 \textbf{0.704} &              \textbf{0.690} &                 \textbf{0.735} &             0.688 &               \textbf{0.607} &                       0.405 &                       \textbf{0.433} &                    \textbf{0.853} &                  \textbf{0.527} &                          \textbf{0.702} &                  \textbf{0.298} &                     \textbf{0.606} &                   \textbf{0.584} \\
                        IMP &                        \textbf{0.278} &                       \textbf{0.272} &                      \textbf{0.590} &                            \textbf{0.442} &                  \textbf{0.832} &                    \textbf{0.337} &                 \textbf{0.693} &              \textbf{0.703} &                 \textbf{0.726} &             0.665 &               \textbf{0.614} &                       \textbf{0.507} &                       \textbf{0.370} &                    \textbf{0.869} &                  \textbf{0.544} &                          \textbf{0.704} &                  0.265 &                     \textbf{0.586} &                   \textbf{0.645} \\

                           Edgepop-A &                        \textbf{0.445} &                       \textbf{0.352} &                      \textbf{0.585} &                            \textbf{0.475} &                  0.807 &                    \textbf{0.464} &                 \textbf{0.705} &              \textbf{0.715} &                 \textbf{0.735} &             \textbf{0.724} &               \textbf{0.642} &                       \textbf{0.476} &                       \textbf{0.466} &                    \textbf{0.880} &                  0.432 &                          \textbf{0.710} &                  \textbf{0.387} &                     \textbf{0.613} &                   \textbf{0.634} \\
                            
                       SNIP-A &                        \textbf{0.241} &                       \textbf{0.255} &                      \textbf{0.614} &                            \textbf{0.414} &                  0.764 &                    \textbf{0.276} &                 \textbf{0.680} &              \textbf{0.686} &                 0.698 &             \textbf{0.734} &               \textbf{0.601} &                      \textbf{0.511} &                       0.310 &                    \textbf{0.861} &                  0.417 &                          0.658 &                  \textbf{0.439} &                     \textbf{0.633} &                   \textbf{0.655} \\
                          
\bottomrule
\end{tabular}
\caption{AUROC of CIFAR10C corruptions (severity 5) on dense and pruned ResNet18 models. We report the best result across sparsity. Bold indicates an improvement compared to the dense model.}
\label{table:cifar10c_5}
\end{table}

\end{document}